# Data set operations to hide decision tree rules

**Dimitris Kalles**[1] and **Vassilios S. Verykios**[1] and **Georgios Feretzakis**[1] and **Athanasios Papagelis**[2]

**Abstract.** This paper focuses on preserving the privacy of sensitive patterns when inducing decision trees. We adopt a record augmentation approach for hiding sensitive classification rules in binary datasets. Such a hiding methodology is preferred over other heuristic solutions like output perturbation or cryptographic techniques - which restrict the usability of the data - since the raw data itself is readily available for public use. We show some key lemmas which are related to the hiding process and we also demonstrate the methodology with an example and an indicative experiment using a prototype hiding tool.

## 1 INTRODUCTION

Privacy preserving data mining [1] is a quite recent research area trying to alleviate the problems stemming from the use of data mining algorithms to the privacy of the data subjects recorded in the data and the information or knowledge hidden in these piles of data. Agrawal and Srinkant [2] were the first to consider the induction of decision trees from anonymized data, which had been adequately corrupted with noise to survive from privacy attacks. The generic strand of knowledge hiding research [3] has led to specific algorithms for hiding classification rules, like, for example, noise addition by a data swapping process [4].

A key target area concerns individual data privacy and aims to protect the individual integrity of database records to prevent the re-identification of individuals or characteristic groups of people from data inference attacks. Another key area is sensitive rule hiding, the subject of this paper, which deals with the protection of sensitive patterns that arise from the application of data mining techniques. Of course, all privacy preservation techniques strive to maintain data information quality.

The main representative of statistical approaches [5] adopts a parsimonious downgrading technique to determine whether the loss of functionality associated with not downgrading the data, is worth the extra confidentiality. Reconstruction techniques involve the redesign of the public dataset [6][7] from the non-sensitive rules produced by algorithms like C4.5 [8] and RIPPER [9]. Perturbation based techniques involve the modification of transactions to support only non-sensitive rules [10], the removal of tuples associated with sensitive rules [11], the suppression of certain attribute values [12] and the redistribution of tuples supporting sensitive patterns so as to maintain the ordering of the rules [13].

In this paper, we propose a series of techniques to efficiently protect the disclosure of sensitive knowledge patterns in classification rule mining. We aim to hide sensitive rules without compromising the information value of the entire dataset. After an expert selects the sensitive rules, we modify class labels at the tree node corresponding to the tail of the sensitive pattern, to eliminate the gain attained by the information metric that caused the splitting. Then, we appropriately set the values of non-class attributes, adding new instances along the path to the root where required, to allow non-sensitive patterns to remain as unaffected as possible. This approach is of great importance as the sanitized data set can be subsequently published and, even, shared with competitors of the data set owner, as can be the case with retail banking [14].

The rest of this paper is structured in 3 sections. Section 2 describes the dataset operations we employ to hide a rule while attempting to minimally affect the decision tree that would have been produced using the modified dataset. Section 3 contains a detailed pseudo-code specification, and early implementation and experimentation details. Section 4 discusses further research issues and concludes the paper.

## 2 THE BASELINE PROBLEM AND A HEURISTIC SOLUTION

Figure 1 shows a baseline problem, which assumes a binary decision tree representation, with binary-valued, symbolic attributes ($X$, $Y$ and $Z$) and binary classes ($C_1$ and $C_2$).

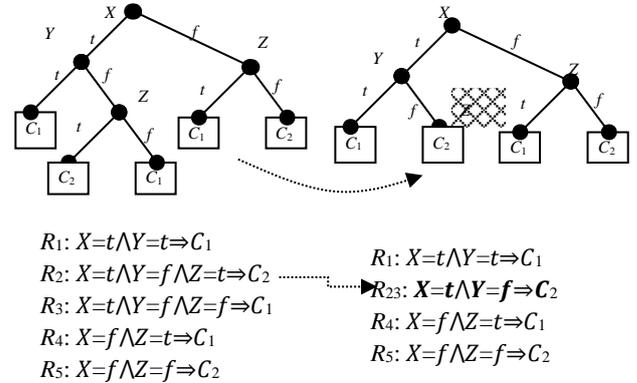

$R_1$: $X=t \wedge Y=t \Rightarrow C_1$
$R_2$: $X=t \wedge Y=f \wedge Z=t \Rightarrow C_2$
$R_3$: $X=t \wedge Y=f \wedge Z=f \Rightarrow C_1$
$R_4$: $X=f \wedge Z=t \Rightarrow C_1$
$R_5$: $X=f \wedge Z=f \Rightarrow C_2$

$R_1$: $X=t \wedge Y=t \Rightarrow C_1$
$R_{23}$: $X=t \wedge Y=f \Rightarrow C_2$
$R_4$: $X=f \wedge Z=t \Rightarrow C_1$
$R_5$: $X=f \wedge Z=f \Rightarrow C_2$

**Figure 1.** A binary decision tree before (left) and after (right) hiding and the associated rule sets.

[1] School of Science and Technology, Hellenic Open University, Patras, Greece,
 email: kalles@eap.gr, verykios@eap.gr, georgios.feretzakis@ac.eap.gr
[2] Epignosis Ltd, Athens, Greece, email: papagel@efrontlearning.net

Hiding $R_3$ implies that the splitting in node $Z$ should be suppressed, hiding $R_2$ as well.

A first idea to hide $R_3$ would be to remove from the training data all the instances of the leaf corresponding to $R_3$ and to retrain the tree from the resulting (reduced) dataset. However this action may incur a substantial tree restructuring, affecting other parts of the tree too.

Another approach would be to turn into a new leaf the direct parent of the $R_3$ leaf. However, this would not modify the actual dataset, thus an adversary could recover the original tree.

To achieve hiding by modifying the original data set in a minimal way, we may interpret "minimal" in terms of changes in the data set or in terms of whether the *sanitized* decision tree produced via hiding is syntactically close to the original one. Measuring minimality in how one modifies decision trees has been studied in terms of heuristics that guarantee or approximate the impact of changes [[15]][[16]][[17]].

However, hiding at $Z$ modifies the statistics along the path from $Z$ to the root. Since splitting along this path depends on these statistics, the relative ranking of the attributes may change, if we run the same induction algorithm on the modified data set. To avoid ending up with a completely different tree, we first employ a bottom-up pass (*Swap-and-Add*) to change the class label of instances at the leaves and then to add some new instances on the path to the root, to preserve the key statistics at the intermediate nodes. Then, we employ a top-down pass (*Allocate-and-Set*) to complete the specification of the newly added instances. These two passes help us hide all sensitive rules and keep the sanitized tree close to the form of the original decision tree.

## 2.1 Adding instances to preserve the class balance

The *Swap-and-Add* pass aims to ensure that node statistics change without threatening the rest of the tree. Using Figure 2a as an example, we show the original tree with class distributions of instances across edges (the original tree does not make use of the parentheses' notation). We use the information gain as the splitting heuristic. To hide the leaf $W_R$ we change the five positive instances to negative ones and denote this operation by $(+5n,-5p)$. As a result the parent node, $W$, becomes a one-class node with minimum (zero) entropy. All nodes located upwards to node $W$ until the root $Z$ also absorb the $(+5n,-5p)$ operation (Figure 2b). At the resulting tree, these primitive initial operations are shown inside parentheses and their results are shown in bold font.

This conversion would leave $X$ with $13n+11p$ instances. But, as its initial $8n+16p$ distribution contributed to $Y$'s splitting attribute, $A_Y$, which in turn created $X$ (and then $W$), we should preserve the information gain of $A_Y$, since the entropy of a node only depends on the ratio $p$:$n$ of its instance classes (Lemma 1).

**Lemma 1** *The entropy of a node only depends on the ratio of its instance classes.*

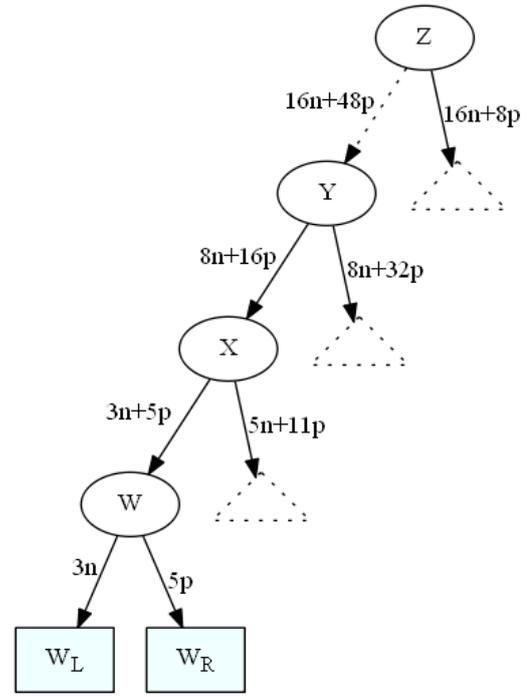

**Figure 2a.** Original tree

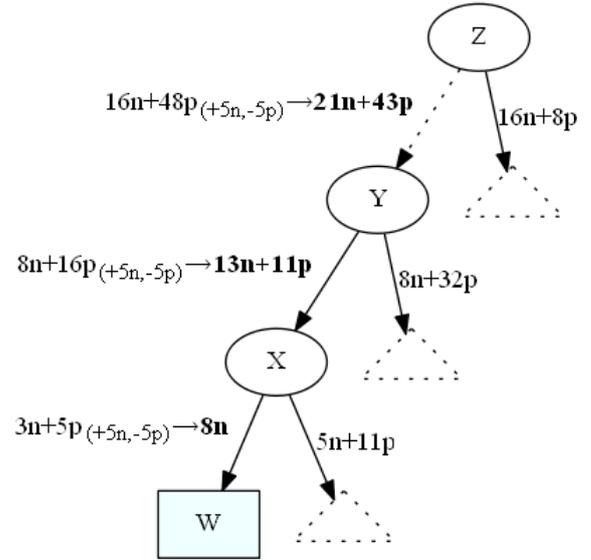

**Figure 2b.** Bottom-up propagation of instances $(+5n,-5p)$

Proof of lemma 1
Let E be the entropy of a node with p positive and n negative instances, with p: n = a. We assume that a ≥ 1.
$p_i$: The probability of the class i

$$E = -\sum p_i \log_2 p_i = -\frac{n}{p+n}\log_2\frac{n}{p+n} - \frac{p}{p+n}\log_2\frac{p}{p+n} =$$

$$\stackrel{p=an}{\cong} -\frac{n}{an+n}\log_2\frac{n}{an+n} - \frac{an}{an+n}\log_2\frac{an}{an+n} =$$

$$= -\frac{1}{a+1}\log_2\frac{1}{a+1} - \frac{a}{a+1}\log_2\frac{a}{a+1} =$$

$$= \frac{1}{a+1}\log_2(a+1) - \frac{a}{a+1}(\log_2 a - \log_2(a+1)) =$$

$$= \log_2(a+1) - \frac{a}{a+1}\log_2 a$$

Now, in the branch YX we have already added some negative instances, while we have also eliminated some positive ones. Ensuring that X's entropy will not increase can be guaranteed by not falling below the 2:1 ratio of one (*any*) class over the other. A greedy option is to add 9*n* instances, to bring the 13:11 ratio to 22:11 (now, 22*n*+11*p* instances arrive at X). These 9*n* instances propagate upwards. To preserve the 3:1 ratio of Y, we must add 47*p* instances (as shown in Figure 3).

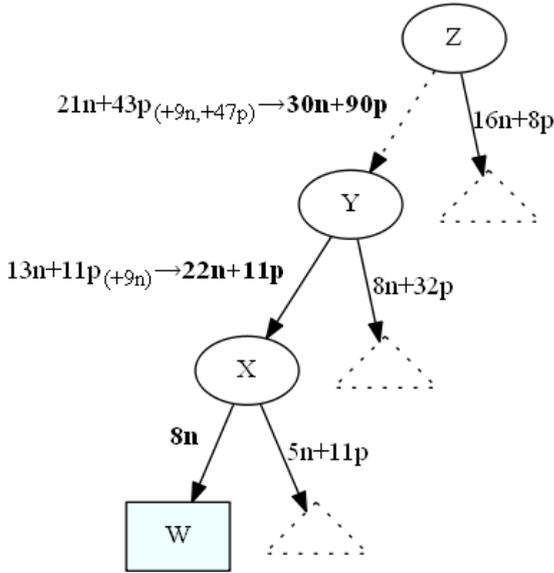

**Figure 3.** A decision tree to demonstrate hiding via instance addition in order to maintain the node statistics, propagating these changes to the tree root.

We extend this technique up until the tree root, by accumulating at each node all instance requests from below and by adding instances locally to maintain the node statistics, propagating these changes to the tree root.

## 2.2 Fully specifying instances

Having set the values of some attributes for the newly added instances is only a partial instance specification, since we have not set those instance values for any other attribute other than the ones present in the path from the root to the node where the instance addition took place. Unspecified values must be so set to ensure that currently selected attributes at all nodes do not get displaced by competing attributes. This is what the *Allocate-and-Set* pass does.

With reference to Figure 3 and the 9*n* instances added due to X via the YX branch, these instances have not had their values set for $A_X$ and $A_W$. Moreover, these must be so set to minimize the possibility that $A_Y$ is displaced from Y, since (at Y) any of attributes $A_W$, $A_X$ or $A_Y$ (or any other) can be selected. Those 9*n* instances were added to help guarantee the existence of X. To simplify things, assume that Y's parent is the root, Z. We will first show how to tackle the top 47*p* instances added and, then, we will build a generic solution.

Since $A_Z$ has been already set for these 47*p* instances, we examine whether these instances will be directed to the YX branch or to the other one (assume that an attribute value of *true* directs an instance to the left branch). As it happened in the bottom-up pass, we need the information gain of $A_Y$ to be large enough to fend off competition from $A_W$ or $A_X$ at node Y, but not too large to threaten $A_Z$. We start with the best possible allocation of values to attribute $A_Y$, directing all 47*p* instances to *false* (toward the 8*n*+32*p* branch), and progressively explore directing some of these along the YX branch, and stop when the information gain for $A_Y$ becomes lower than the information gain for $A_Z$. We use the term *two-level hold-back* to refer to this technique, as it spans two tree levels. This approach exploits the convexity property of the information gain difference function (Lemma 2).

The *Allocate-and-Set* pass examines all four combinations of distributing all positive and all negative instances to one branch, select the one that maximizes the information gain difference and then move along the slope that decreases the information gain, until we do not exceed the information gain of the parent; then perform the recursive specification all the way to the tree fringe.

**Lemma 2** *Distributing new class instances along only one branch maximizes information gain.*

Proof of lemma 2

Let $G(i)$ be the function that represents the information gain after the addition of k new positive instances at a node and the distribution of $i$ of these nodes to the left child and of $k - i$ to the right child.

$$G(i) = G(p+k, n, p_1+i, n_1, p_2+(k-i), n_2), for\ 0 \le i \le k$$

$$G(i) = E(p+k, n) - \left[E(p_1+i, n_1)\cdot\frac{p_1+i+n_1}{p+k+n} + E(p_2+(k-i), n_2)\cdot\frac{p_2+k-i+n_2}{p+k+n}\right] =$$

$$= \overbrace{E(p+k, n)}^{C-Constant} - \left[\left(\log\frac{p_1+i+n_1}{n_1} - \frac{p_1+i}{p_1+i+n_1}\cdot\log\frac{p_1+i}{n_1}\right)\cdot\frac{p_1+i+n_1}{p+k+n}\right.$$
$$+ \left(\log\frac{p_2+k-i+n_2}{n_2} - \frac{p_2+k-i}{p_2+k-i+n_2}\right.$$
$$\left.\left.\cdot\log\frac{p_2+k-i}{n_2}\right)\cdot\frac{p_2+k-i+n_2}{p+k+n}\right] =$$

$$= C - \frac{1}{p+k+n}\left[(p_1+i+n_1)\cdot\log\frac{p_1+i+n_1}{n_1} - (p_1+i)\cdot\log\frac{p_1+i}{n_1}\right.$$
$$+ (p_2+k-i+n_2)\cdot\log\frac{p_2+k-i+n_2}{n_2}$$
$$\left.- (p_2+k-i)\cdot\log\frac{p_2+k-i}{n_2}\right]$$

$$\stackrel{\text{change of base}}{\triangleq} C - \frac{1}{p+k+n}$$
$$\cdot\frac{1}{\ln 2}\left[(p_1+i+n_1)\cdot\ln\frac{p_1+i+n_1}{n_1} - (p_1+i)\right.$$
$$\cdot\ln\frac{p_1+i}{n_1} + (p_2+k-i+n_2)\cdot\ln\frac{p_2+k-i+n_2}{n_2}$$
$$\left.- (p_2+k-i)\cdot\ln\frac{p_2+k-i}{n_2}\right]$$

Taking the first derivative of $G(i)$ we have:

$$G'(i) = 0 - \frac{1}{p+k+n}$$
$$\cdot\frac{1}{\ln 2}\left[\ln\frac{p_1+i+n_1}{n_1} + (p_1+i+n_1)\cdot\frac{n_1}{p_1+i+n_1}\right.$$
$$\cdot\frac{1}{n_1} - \ln\frac{p_1+i}{n_1} - (p_1+i)\cdot\frac{n_1}{p_1+i}\cdot\frac{1}{n_1}$$
$$+ (-1)\cdot\ln\frac{p_2+k-i+n_2}{n_2} - (p_2+k-i+n_2)$$
$$\cdot\frac{n_2}{p_2+k-i+n_2}\cdot\frac{1}{n_2} - (-1)\cdot\ln\frac{p_2+k-i}{n_2}$$
$$\left.- (p_2+k-i)\cdot\frac{n_2}{p_2+k-i}\cdot\left(-\frac{1}{n_2}\right)\right] =$$

$$= -\frac{1}{p+k+n}\left[\ln\frac{p_1+i+n_1}{n_1} + 1 - \ln\frac{p_1+i}{n_1} - 1\right.$$
$$\left.- \ln\frac{p_2+k-i+n_2}{n_2} - 1 + \ln\frac{p_2+k-i}{n_2} + 1\right] =$$

$$= -\frac{1}{p+k+n}\left[\ln\frac{p_1+i+n_1}{p_1+i} + \ln\frac{p_2+k-i}{p_2+k-i+n_2}\right] =$$

$$= -\frac{1}{p+k+n}\left[\ln\frac{(p_1+i+n_1)\cdot(p_2+k-i)}{(p_1+i)\cdot(p_2+k-i+n_2)}\right]$$

Next we find the second derivative function of $G(i)$

$$G''(i) = -\frac{1}{p+k+n}\left[\ln\frac{(p_1+i+n_1)\cdot(p_2+k-i)}{(p_1+i)\cdot(p_2+k-i+n_2)}\right]' =$$

$$= -\frac{1}{p+k+n}\cdot\underbrace{\frac{(p_1+i)\cdot\overbrace{(p_2+k-i+n_2)}^{\geq 0}}{(p_1+i+n_1)\cdot\underbrace{(p_2+k-i)}_{\geq 0}}}_{A(i)<0 \text{ for every } i\in[0,k]}$$
$$\cdot\left(\frac{(p_1+i+n_1)\cdot(p_2+k-i)}{(p_1+i)\cdot(p_2+k-i+n_2)}\right)' =$$

$$= A(i)\cdot\frac{((p_1+i+n_1)\cdot(p_2+k-i))'\cdot((p_1+i)\cdot(p_2+k-i+n_2))}{\underbrace{((p_1+i)\cdot(p_2+k-i+n_2))^2}_{B(i)>0}}$$
$$- \frac{((p_1+i+n_1)\cdot(p_2+k-i))\cdot((p_1+i)\cdot(p_2+k-i+n_2))'}{\underbrace{((p_1+i)\cdot(p_2+k-i+n_2))^2}_{B(i)>0}} =$$

$$= \overbrace{\left(\frac{A(i)}{B(i)}\right)}^{C(i)<0}\cdot[(p_2+k-i-p_1-i-n_1)\cdot(p_1+i)\cdot(p_2+k-i+n_2)$$
$$- (p_1+i+n_1)\cdot(p_2+k-i)$$
$$\cdot(p_2+k-i-p_1-i+n_2)] =$$

$$= C(i)\cdot[((p_2+k-i)-(p_1+i+n_1))\cdot(p_1+i)\cdot(p_2+k-i+n_2)$$
$$- (p_1+i+n_1)\cdot(p_2+k-i)$$
$$\cdot((p_2+k-i+n_2)-(p_1+i))] =$$

$$= C(i)\cdot[(p_2+k-i)\cdot(p_1+i)\cdot(p_2+k-i+n_2) - (p_1+i+n_1)$$
$$\cdot(p_1+i)\cdot(p_2+k-i+n_2) - (p_2+k-i+n_2)$$
$$\cdot(p_1+i+n_1)\cdot(p_2+k-i) + (p_1+i)$$
$$\cdot(p_1+i+n_1)\cdot(p_2+k-i)] =$$

$$= C(i)\cdot[(p_2+k-i)\cdot(p_2+k-i+n_2)(p_1+i-p_1-i-n_1)$$
$$+ (p_1+i)\cdot(p_1+i+n_1)$$
$$\cdot(p_2+k-i-p_2-k+i-n_2)] =$$

$$= C(i)\cdot[(p_2+k-i)\cdot(p_2+k-i+n_2)\cdot(-n_1) + (p_1+i)$$
$$\cdot(p_1+i+n_1)\cdot(-n_2)] =$$

$$= -C(i)\cdot$$
$$\underbrace{\left[n_1\cdot\overbrace{(p_2+k-i)}^{\geq 0}\cdot\overbrace{(p_2+k-i+n_2)}^{\geq 0} + n_2\cdot(p_1+i)\cdot(p_1+i+n_1)\right]}_{D(i)>0} =$$

$$= -C(i)\cdot D(i) > 0, \text{for every } i\in[0,k]$$

So, the function $G(i)$ is convex in the interval $[0,k]$.

The proof of the Lemma 3 is completed due to a theorem that states *"A convex function on a closed bounded interval attains its maximum at one of its endpoints"*.

## 2.3    Grouping of hiding requests

By serially processing hiding requests, each one incurs the full cost of updating the instance population. By knowing all of them in advance, we only consider once each node in the bottom-up pass and once in the top-down pass. We express that dealing with all hiding requests in parallel leads to the minimum number of new instances by:

$$\left|T_R^P\right| = \min_i\left|\left(T_{\{i\}}^S\right)_{R-\{i\}}^S\right|$$

The formula states that for a tree $T$, the number of instances ($|T|$), after a parallel ($T^p$) hiding process of all rules (leaves) in $R$, is the optimal along all possible orderings of all serial ($T^s$) hiding requests drawn from $R$. A serial hiding request is implemented by selecting a leaf to be hidden and then, recursively, dealing with the remaining leaves (Lemma 3). As an exception, the *max:min* ratio is not affected by simultaneously requesting to hide two sibling pure-class leaves.

**Lemma 3** *When serially hiding two non-sibling leaves, the number of new instances to be added to maintain the* max:min *ratios is larger or equal to the number of instances that would have been added if the hiding requests were handled in parallel.*

Proof sketch

Let a parent node have p positive and n negative instances, with $p:n = r \geq 1$, and let two hiding requests having propagated up to that node, each from a different branch, demanding that $p_L, n_L$ (from the left child) and $p_R, n_R$ (from the right child) instances be respectively added. Assume now, that to maintain the $p:n$ ratio, we need to add at once (parallel) $p_X$ or $n_X$ instances to parent node instead of adding first $p_1$ or $n_1$ (in order to control the change due to left child) and then adding $p_2$ or $n_2$ (in order to control the change due to right child). First, we construct a table (Table 1) with all (32) possible cases and then treat each combination as a separate case.

As example we present the following proofs of eight possible cases:

**Case (I-1):**

$$\frac{a+p_{X_I}}{b} = \frac{a+p_1+p_2}{b} \Leftrightarrow p_{X_I} = p_1 + p_2 \quad Q.E.D.$$

**Case (I-2):**

$$\frac{a+p_{X_I}}{b} = \frac{b}{a+p_1+p_2}$$

The reason that we select the option $(I), i.e. \frac{a+p_{X_I}}{b}$, is that $p_{X_I}$ is the minimum number of instances to add in order to maintain the ratio in the parent node. Therefore, $p_{X_I} = min\{p_{X_I}, p_{X_{II}}, n_{X_{III}}, n_{X_{IV}}\}$ which means that $p_{X_{II}} \geq p_{X_I}$ (∗). If we had selected the option $(II), i.e. \frac{b}{a+p_{X_{II}}}$ we would have the case (II-2).

i.e. $\frac{b}{a+p_{X_{II}}} = \frac{b}{a+p_1+p_2} \Leftrightarrow p_{X_{II}} = p_1 + p_2$ (∗∗)

From (∗), (∗∗) we have $p_1 + p_2 \geq p_{X_I}$ Q.E.D.

**Case (I-3):**

$$\frac{a+p_{X_I}}{b} = \frac{a+p_1}{b+n_2} \Leftrightarrow$$
$$\Leftrightarrow ab + an_2 + bp_{X_I} + n_2 p_{X_I} = ab + bp_1 \Leftrightarrow$$
$$\Leftrightarrow an_2 + b(p_{X_I} - p_1) + p_{X_I} n_2 = 0$$

- If $p_{X_I} \geq p_1$ then this case is impossible, since all the terms in the left hand side in the above equation are positive.
- If $p_{X_I} < p_1$ then we have $p_1 + n_2 > p_{X_I}$ Q.E.D.

**Case (I-4):**

$$\frac{a+p_{X_I}}{b} = \frac{b+n_2}{a+p_1}$$

The reason that we select the option $(I), i.e. \frac{a+p_{X_I}}{b}$, is that $p_{X_I}$ is the minimum number of instances to add in order to maintain the ratio in the parent node. Therefore, $p_{X_I} = min\{p_{X_I}, p_{X_{II}}, n_{X_{III}}, n_{X_{IV}}\}$ which means that $p_{X_{II}} \geq p_{X_I}$ (∗). If we had selected the option $(II), i.e. \frac{b}{a+p_{X_{II}}}$ we would have the case (II-4) and for that it had been proved that

$$p_1 + n_2 > p_{X_{II}} \quad (**)$$

From (∗), (∗∗) we have $p_1 + n_2 > p_{X_I}$ Q.E.D.

**Case (I-5):**

$$\frac{a+p_{X_I}}{b} = \frac{a}{b+n_1+n_2} \Leftrightarrow$$
$$\Leftrightarrow ab + a(n_1+n_2) + bp_{X_I} + p_{X_I}(n_1+n_2) = ab \Leftrightarrow$$
$$\Leftrightarrow a(n_1+n_2) + bp_{X_I} + p_{X_I}(n_1+n_2) = 0$$

This case is impossible, since all the terms in the left hand side in the above equation are positive.

**Case (I-6):**

$$\frac{a+p_{X_I}}{b} = \frac{b+n_1+n_2}{a}$$

The reason that we select the option $(I), i.e. \frac{a+p_{X_I}}{b}$, is that $p_{X_I}$ is the minimum number of instances to add in order to maintain the ratio in the parent node. Therefore, $p_{X_I} = min\{p_{X_I}, p_{X_{II}}, n_{X_{III}}, n_{X_{IV}}\}$ which means that $n_{X_{IV}} \geq p_{X_I}$ (∗). If we had selected the option $(IV), i.e. \frac{b+n_{X_{IV}}}{a}$ we would have the case (IV-6),

i.e. $\frac{b+n_{X_{IV}}}{a} = \frac{b+n_1+n_2}{a} \Leftrightarrow n_{X_{IV}} = n_1 + n_2$ (∗∗)

From (∗), (∗∗) we have $n_1 + n_2 > p_{X_I}$ Q.E.D.

**Case (I-7):**

$$\frac{a+p_{X_I}}{b} = \frac{a+p_2}{b+n_1} \Leftrightarrow$$
$$\Leftrightarrow ab + an_1 + bp_{X_I} + n_1 p_{X_I} = ab + bp_2 \Leftrightarrow$$
$$\Leftrightarrow an_1 + b(p_{X_I} - p_2) + n_1 p_{X_I} = 0$$

- If $p_{X_I} \geq p_2$ then this case is impossible, since the left hand side in the above equation is positive.
- If $p_{X_I} < p_2$ then we have $p_2 + n_1 > p_{X_I}$ Q.E.D.

**Case (I-8):**

$$\frac{a+p_{X_I}}{b} = \frac{b+n_1}{a+p_2}$$

The reason that we select the option $(I), i.e. \frac{a+p_{X_I}}{b}$, is that $p_{X_I}$ is the minimum number of instances to add in order to maintain the ratio in the parent node. Therefore, $p_{X_I} = min\{p_{X_I}, p_{X_{II}}, n_{X_{III}}, n_{X_{IV}}\}$ which means that $p_{X_{II}} \geq p_{X_I}$ (∗). If we had selected the option $(II), i.e. \frac{b}{a+p_{X_{II}}}$ we would have the case (II-8) and for that it had been proved that

$$n_1 + p_2 > p_{X_{II}} \quad (**)$$

From $(*)$, $(**)$ we have $n_1 + p_2 > p_{X_I}$   Q.E.D.

The proofs of all other cases are similar (omitted due to space limitations).

**Table 1**: All possible cases to compare serially and parallel addition of new instances.

| Parallel | Serially |
|---|---|
| $\dfrac{a + p_{X_I}}{b}$   (I) | $\dfrac{a + p_1 + p_2}{b}$   (1) |
| | $\dfrac{b}{a + p_1 + p_2}$   (2) |
| $\dfrac{b}{a + p_{X_{II}}}$   (II) | $\dfrac{a + p_1}{b + n_2}$   (3) |
| | $\dfrac{b + n_2}{a + p_1}$   (4) |
| $\dfrac{a}{b + n_{X_{III}}}$   (III) | $\dfrac{a}{b + n_1 + n_2}$   (5) |
| | $\dfrac{b + n_1 + n_2}{a}$   (6) |
| $\dfrac{b + n_{X_{IV}}}{a}$   (IV) | $\dfrac{a + p_2}{b + n_1}$   (7) |
| | $\dfrac{b + n_1}{a + p_2}$   (8) |

## 3   A PROTOTYPE IMPLEMENTATION

A prototype system allows a user to experiment with our hiding heuristic in binary-class, binary-value data sets. For a brief experiment we used a home-grown data generator; we generated 1,000 instances for a 5-attribute problem and distributed those instances uniformly over 11 rules (Table 2).

**Table 2.** Rule notation: $(t,\_,\_,f,\_)$ means "*if* $(A_1 = t)$ & $(A_4 = f)$ *then* ...".

```
(t,t,t,t,t):p    (t,t,f,t,_):p
(t,t,t,t,f):n    (t,t,f,f,_):n
(t,t,t,f,t):p
(t,t,t,f,f):n
```

```
(t,f,t, _,_):p   (t,t,_, _,_):p
(t,f,f, _,_):n
(f,f,f, _,_):n
(f,f,t,_,_):n
```

For the original decision tree we then observed, for each leaf node, the number of instances we would need to add to hide just that node. The average increase is about 67% with deep fringe nodes (longer rules) generating relatively light changes and shallow fringe nodes (shorter rules) generating larger ones.

A further experiment highlighted that large increases occur when we want to hide eminent rules. For example, when we tested our technique with a modified version of the rule set in Table 2, one that was produced by removing all rules of length 3 and 4, it turned out that the average increase over all leaves was about 73%, with some leaves accounting for a nearly 400% increase and some others for a mere 10%. When we removed all rules of length 2 and 3, the average increase was 80%. As short, eminent, rules involve fewer attributes, skewing the statistics for these attributes entails a substantial dataset modification.

Large numbers of instances to be added do not mean that the tree structure will also change a lot; we usually succeed to keep the form of the sanitized tree as close as possible to the original one. Still, the growth ratio can be quite large and this motivates the grouping of hiding requests.

## 4   CONCLUSIONS AND DIRECTIONS FOR FURTHER WORK

We have presented the outline of a heuristic that allows one to specify which leaves of a decision tree should be hidden and then proceed to judiciously add instances to the original data set so that the next time one tries to build the tree, the to-be-hidden nodes will have disappeared because the instances corresponding to those nodes will have been absorbed by neighboring ones.

We have presented a fully-fledged example of the proposed approach and, along its presentation, discussed a variety of issues that relate to how one might minimize the amount of modifications that are required to perform the requested hiding as well as where some side-effects of this hiding might emerge. We have also presented the heuristic in pseudo-code, mentioned the development of a prototype system (open to all) and demonstrated its use in synthetic data alongside some explanations for observed results.

Of course, several aspects of our technique can be substantially improved.

The instance adding scheme is greedy. For a simple example, refer to Figure 3, where we added $9n$ instances along the *YX* branch and generated a need for $47p$ instances at the top. Had we added $15p$ instances instead to maintain the 2:1 ratio ($13n{:}26p$), we would have ended up in a 21:58 ratio at the upper branch, which would only require a further $5p$ instances to maintain its original 3:1 ratio; we would need 20 instances instead of 56. We are now developing a full look-ahead technique based on linear Diophantine equations.

The *max:min* ratio concept can guarantee the preservation of the information gain of a splitting attribute but it would be interesting to see whether it can be applied to other splitting criteria too. Since this ratio is based on frequencies, it should also work with a similar popular metric, the *Gini* index [[18]]. On the other hand, it is unclear whether it can preserve trees that have been induced using more holistic metrics, such as the minimum description length principle [[19]].

Extensive experimentation with several data sets would allow us to estimate the quality of the *max:min* ratio heuristic and also experiment with a revised version of the heuristic, one that strives to keep the *p:n* ratio of a node itself (and not its parent), or one that attempts to remove instances instead of swapping their class labels, or still another that further relaxes the *p:n* ratio concept during the top-down phase by distributing all unspecified instances evenly among the left and right outgoing branch from a node and proceeding recursively to the leaves (which is the one we actually implemented). In general, experimenting with a variety of heuristics to trade off ease of implementation with performance is an obvious priority for experimental research.

On performance aspects, besides speed, one also needs to look at the issue of judging the similarity of the original tree with the one produced after the above procedure has been applied. One might be interested in syntactic similarity [[20]] (comparing the data structures –or parts thereof- themselves) or semantic similarity (comparing against reference data sets). This is an issue of substantial importance, which will also help settle questions of which heuristics work better and which not.

As an extreme example of the need to experiment exhaustively with the proposed heuristic, consider a simple implementation of the hiding primitive introduced in Figure 2a: instead of swapping the class label of a leaf, opt to swap the value of the test attribute at node *W*; this also eliminates *W*'s contribution to *X* and renders the existence of *X* questionable in a re-run of the induction algorithm. Our heuristic aims at minimizing the impact on the original data set but since such impact will have to be weighed against other problem parameters, the issue is open for investigation.

It is rather obvious that the variety of answers one could explore for each of the questions above constitutes a research agenda of both a theoretical and an applied nature. At the same time, it is via extending the base case, by allowing multi-valued and numeric attributes and multi-class problems that we should address the problem of enhancing our basic technique, alongside investigating the robustness of our heuristic to a variety of splitting criteria and to datasets of varying size and complexity. The longer-term goal is to have it operate as a standard data engineering service to accommodate hiding requests, coupled with a suitable environment where one could specify the importance of each hiding request. The current prototype (available at http://www.splendor.gr/trees) serves exactly to highlight this goal.